\documentclass{article} 
\usepackage{iclr2026_conference,times}

\usepackage{amsmath,amsfonts,bm}

\def\eqref#1{equation~\ref{#1}}

\def\1{\bm{1}}

\DeclareMathAlphabet{\mathsfit}{\encodingdefault}{\sfdefault}{m}{sl}
\SetMathAlphabet{\mathsfit}{bold}{\encodingdefault}{\sfdefault}{bx}{n}

\usepackage{hyperref}
\usepackage{url}
\usepackage{comment}
\usepackage{amsmath}
\usepackage{booktabs}
\usepackage{graphicx}
\usepackage[utf8]{inputenc}
\usepackage{amsthm}
\newtheorem{theorem}{Theorem}
\usepackage{booktabs,tabularx,threeparttable,siunitx,makecell}
\newcolumntype{d}[1]{S[table-format=#1,table-number-alignment=center,table-column-width=1.45cm]}

\title{AdaProb: Efficient Machine Unlearning via Adaptive Probability}

\author{Zihao Zhao\\
Johns Hopkins University\\
\texttt{zzhao71@jhu.edu} \\
\And
Yuchen Yang \\
The Pennsylvania State University \\
\texttt{yuchen.yang@psu.edu} \\
\And
Anjalie Field \\
Johns Hopkins University \\
\texttt{afield6@jhu.edu}
\And
Yinzhi Cao \\
Johns Hopkins University \\
\texttt{yinzhi.cao@jhu.edu}
}

\iclrfinalcopy 
\begin{document}

\maketitle
\begin{abstract}
Machine unlearning, enabling a trained model to forget specific data, is crucial for addressing erroneous data and adhering to privacy regulations like the General Data Protection Regulation (GDPR)'s ``right to be forgotten". Despite recent progress, existing methods face two key challenges: residual information may persist in the model even after unlearning, and the computational overhead required for effective data removal is often high. To address these issues, we propose Adaptive Probability Approximate Unlearning (AdaProb), a novel method that enables models to forget data efficiently and in a privacy-preserving manner. Our method firstly replaces the neural network's final-layer output probabilities with pseudo-probabilities for data to be forgotten. These pseudo-probabilities follow a uniform distribution to maximize unlearning, and they are optimized to align with the model’s overall distribution to enhance privacy and reduce the risk of membership inference attacks. Then, the model's weights are updated accordingly. Through comprehensive experiments, our method outperforms state-of-the-art approaches with over 20\% improvement in forgetting error, better protection against membership inference attacks, and less than 50\% of the computational time.\footnote{The code is provided in https://github.com/zzhao71/AdaProb.git}
\end{abstract}
\section{Introduction}

Machine unlearning focuses on eliminating the impact of specific data subsets, such as erroneous, or privacy-leaking instances~\citep{jagielski2018manipulating, yang2023sneakyprompt} used in model training~\citep{baumhauer2022machine,fu2022knowledge,golatkar2020eternal,golatkar2020forgetting,guo2019certified,kim2022efficient,mehta2022deep,nguyen2020variational,shah2023unlearning}. It has emerged as a critical area of research due to growing concerns about data privacy \citep{pardau2018california}, legal requirements for data deletion \citep{mantelero2013eu}, and the need for models to adapt to new information without complete retraining. Though the most straightforward approach is to retrain the model with a new dataset that excludes the data needing removal, this approach is computationally expensive and needs continuous access to the entire training set.

One of the most prominent use cases for machine unlearning is privacy protection.~\citep{nguyen2022survey}.
In this case, unlearning aims to modify the model to forget a set of training data points, so that an adversary cannot determine anything about them from the model, including whether or not they were part of the training set~\citep{hu2024learn}.
Conventional unlearning methods often fail to achieve this behavior: in forcing the model to perform poorly on the forget set (i.e., exhibit high loss), they create a distinguishable pattern between forgotten and retained data ~\citep{wang2024llm, chen2021machine}. This performance disparity enables attackers to identify forgotten samples through techniques like Membership Inference Attacks (MIA)~\citep{BlindMI}. To prevent this vulnerability, effective privacy-preserving unlearning must ensure
the model's behavior on forgotten data is indistinguishable from its behavior had it never encountered that data during training ~\citep{guo2019certified, xu2023machine}. This often requires sacrificing some model performance to achieve stronger privacy protection ~\citep{qu2023learn}. 
\begin{figure}[t]  
\includegraphics[width=1.0\textwidth]{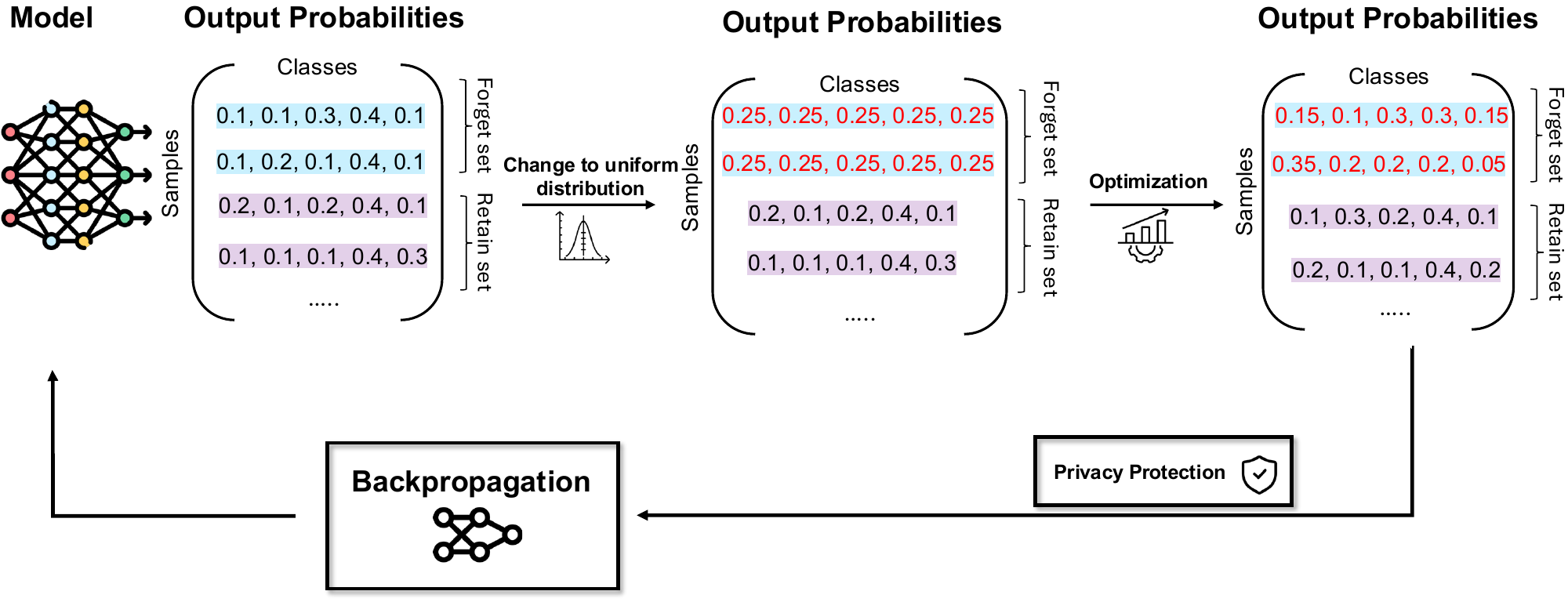}  
    \caption{This is an overview of Adaptive Probability Unlearning (AdaProb). In this approach, we extract the output layer probabilities and replace the forget set probabilities with pseudo-probabilities. After performing optimization, the model's weights are fine-tuned using the refined pseudo-probabilities.}
    \label{overview_fig}
\end{figure}

Current machine unlearning methods primarily focus on either gradient-based approaches ~\citep{neel2021descent} that optimize modified loss functions to induce forgetting or adjusting the model architecture through layer addition/deletion. We propose a fundamentally different approach that manipulates the final-layer output probabilities and leverages backpropagation to update model weights accordingly.

Our Adaptive Probability Approximate Unlearning (AdaProb) method replaces the model's output probabilities with uniformly distributed pseudo-probabilities for forget-set data, ensuring effective forgetting. Then, we iteratively refine the output probability distribution to align forget-set outputs with pseudo-probabilities while constraining retain-set outputs to remain similar to the original model's predictions. This dual constraint ensures forgotten samples become indistinguishable from retained data, preventing information leakage. The overview of the method is illustrated in Figure~\ref{overview_fig}. 

Our extensive evaluations demonstrate that AdaProb achieves a 50\% reduction in computational time compared to state-of-the-art methods while simultaneously improving unlearning effectiveness. Notably, AdaProb reduces the success rate of membership inference attacks to near-random guessing levels, validating its strong privacy protection capabilities.

Our contributions are threefold: 
(1) We propose a novel unlearning method based on output probability manipulation that handles privacy tasks. 
(2) We achieve 50\% computational speedup over existing methods while improving unlearning performance.
(3) We provide comprehensive experimental validation demonstrating superior privacy protection, with membership inference attacks reduced to random-guessing levels.

\section{Related work}

\paragraph{Machine Unlearning}
Machine unlearning, first proposed by~\citet{cao2015towards}, has evolved into two main paradigms: exact unlearning, which ensures complete data removal, and approximate unlearning, which reduces data influence to acceptable levels~\citep{izzo2021approximate}. While exact unlearning methods have been developed for specific models~\citep{brophy2021machine, schelter2021hedgecut, ginart2019making}, they suffer from prohibitive computational costs, especially as model size increases. 

Approximate unlearning methods have been developed to address the computational challenges of high-dimensional neural networks. These approaches employ various strategies: weight modification methods directly adjust model parameters~\citep{golatkar2020eternal, golatkar2020forgetting}, architectural approaches like SISA training partition data during training to facilitate removal~\citep{bourtoule2021machine}, while others leverage cached gradients~\citep{wu2020deltagrad} or optimization techniques~\citep{kurmanji2024towards} to accelerate retraining. Additionally, certified unlearning methods formalize the unlearning goal by requiring the unlearned model to be provably close to one retrained from scratch on only the retained data~\citep{zhang2024towards}. However, these approaches face critical limitations: they are still computationally expensive, often compromise model utility, and, most importantly, fail to address privacy protection against membership inference attacks perfectly.

This gap motivates our approach: rather than modifying parameters directly or restructuring training, we manipulate output probabilities to achieve efficient unlearning while providing strong privacy guarantees. Our method addresses the key shortcomings of existing work by reducing computational time by 50\% and explicitly protecting against privacy attacks, achieving near-random membership inference success rates.
\paragraph{Machine Unlearning Evaluations}To evaluate unlearning methods, it is common to compare models before and after unlearning across three key metrics: computational efficiency, model utility, and privacy protection. Computational efficiency is measured by the time required for the unlearning process, while model utility is assessed by comparing test set performance before and after unlearning. Privacy protection, however, is more challenging to measure. Current approaches include: (1) comparing posterior distributions or parameters between retrained and unlearned models \citep{golatkar2020eternal,golatkar2020forgetting}, (2) providing theoretical guarantees or bounds for the unlearned model \citep{chien2022efficient, guo2019certified, neel2021descent}, and (3) applying attacks to measure privacy risks \citep{chen2021machine}, such as membership inference attacks (MIA)~\citep{shokri2017membership}, which use shadow models to generate synthetic training data for attack classifiers. In the paper, we measure the privacy protection through membership inference attack, and the KL divergence of output distribution on the forget set between the retrained model and the unlearned model.

\section{Notations and Problem Definition}
Consider a dataset ${(\mathbf{x}, y)} \in \mathcal{D}$, composed of $N$ data points, where each instance consists of an input feature vector $\mathbf{x}_i$ and its corresponding label $y_i$. Let $f(\cdot; \mathbf{w})$ represent a function implemented by a deep neural network, parameterized by the weights $\mathbf{w}$. In this context, we are provided with a ``forget set" $\mathcal{D}_\text{f} = {(\mathbf{x}_i, y_i)}_{i=1}^{N_f} \subset \mathcal{D}$, consisting of $N_f$ instances extracted from $\mathcal{D}$, as well as a``retain set" $\mathcal{D}_\text{r} = {(\mathbf{x}_j, y_j)}_{j=1}^{N_r} \subset \mathcal{D}$ containing $N_r$ training samples. For simplicity, we assume that $\mathcal{D}_\text{r}$ is the complement of $\mathcal{D}_\text{f}$, satisfying the condition $\mathcal{D}_\text{f} \cup \mathcal{D}_\text{r} = \mathcal{D}$ and $N_f + N_r = N$, thereby covering the entire original dataset.

The goal of machine unlearning is to derive a new set of weights, $\mathbf{w}_u$, such that the updated model, $f(\cdot; \mathbf{w}_u)$, effectively forget the information related to $\mathcal{D}_\text{f}$. Specifically, the unlearned model should maintain its original performance on the retain set $D_\text{r}$ and retain its ability to generalize to unseen data. In the paper, we use ``original'' to indicate the pre-unlearning model.

\section{Methods}
Building on the foundational framework, we propose a machine unlearning approach that optimizes the output probabilities at the final layer and subsequently backpropagates these adjustments to update the model weights throughout the network. 

We define the output layer probability distribution for each data point as a $k$-dimensional vector, where $k$ is the number of classes. Let $f(x,\mathbf{w})$ denote the output probability distribution generated when input $x$ is passed through the model with weights $\mathbf{w}$. For the forget set $D_\text{f}$ and retain set $D_\text{r}$, we denote their output distributions as $\{f(x_i;\mathbf{w}) \}_{i=1}^{N_f}$ and $\{f(x_j;\mathbf{w}) \}_{j=1}^{N_r}$, respectively.


The core of our method lies in formulating an optimization objective that adjusts the model's output distribution to effectively forget the information in the forget set $\mathcal{D}_\text{f}$ while preserving performance on the retain set $\mathcal{D}_\text{r}$. We first change the forget set output distributions to uniform distributions to obscure learned patterns, and keep the original model output distribution for the retain set to maintain performance. Then, the optimization minimizes the discrepancy between current outputs $g(x;\mathbf{w})$ and original distributions $f(x;\mathbf{w})$. After obtaining the optimal output distributions, we backpropagate to update the model weights, teaching the network to realize these adjusted predictions.


\subsection{Pseudo-Probability Refinement}

To find the optimal output distributions, we start by replacing the model's output distribution with a pseudo-probabilistic distribution, such as a uniform distribution. The rationale behind this strategy is to ``mask" or obscure the model's learned associations with the forget set by assigning equal probabilities to each class, thereby eliminating the model's ability to make confident predictions on these data points. 


Specifically, we construct a probability matrix, where each row represents an input data point and each column represents a class. For a dataset with $N$ data points and $K$ classes, the matrix has dimension $N\times K$. Each element $(i,k)$ contains $g_k(x_i;\mathbf{w})$, the probability of data point $x_i$ belonging to class $k$. The matrix representation facilitates our optimization constraints on both row sums and column sums. 

In our formulation, $\{g(x_i,\mathbf{w})\}_{i=1}^{N_f}$  denotes the uniform pseudo-probability distribution for the forget set $D_\text{f}$. These are designed to disrupt the model's learned patterns while preserving performance on the retain set. For the retain set, $\{g(x_j,\mathbf{w})\}_{j=1}^{N_r}$  is set to the original model outputs $\{f(x_j,\mathbf{w_\text{original}})\}_{j=1}^{N_r}$. Given a data point $x_i$ in the forget set, $g_k(x_i,\mathbf{w})$ denotes its probability of belonging to class $k$, where $k \in \{1, ..., K\}$



In the privacy setting, directly using uniform distributions makes the model vulnerable to membership inference attacks, as such artificial patterns are easily detectable. We apply our optimization to minimize the KL divergence between current and original distributions for both sets, balanced by parameter $\lambda$. This can let the forget set output distribution to be similar to the original distribution, which makes it hard to be detected by membership inference attack. 

To address this, we introduce constraints on our probability matrix:  
(1) \textbf{Column constraints:} The sum of each column (total probability for class $k$ across all data points) must equal $M_k = \sum_{i=1}^{N_f}g_k(x_i;\mathbf{w}) + \sum_{j=1}^{N_r}g_k(x_j;\mathbf{w})$
(2) \textbf{Row constraints:} Each row must sum to 1, ensuring valid probability distributions.
(3) \textbf{Element constraints:} All probabilities must lie in $[0,1]$.

The optimization updates the output distribution  $\{g(x_i,\mathbf{w})\}_{i=1}^{N_f}$ for the forget set and $\{g(x_j,\mathbf{w})\}_{j=1}^{N_r}$ for the retain set to find the optimal values that minimize the objective function while satisfying all constraints. 
{\fontsize{8}{9}\selectfont 
  \setlength{\jot}{2pt}
\begin{align}
\label{1}
\min_{\{g(x_i; \mathbf{w})\}_{i=1}^{N_f}, \{g(x_j; \mathbf{w})\}_{j=1}^{N_r}} 
& \left( \sum_{i=1}^{N_f} D_{KL}( g(x_i,\mathbf{w}) \|f(x_i; \mathbf{w})) 
 + \lambda \sum_{j=1}^{N_r} D_{KL}(g(x_i,\mathbf{w}) \|f(x_j; \mathbf{w})  )\right) \\
\label{2}
\text{subject to}
\quad & \sum_{i=1}^{N_f} g_k(x_i; \mathbf{w}) + \sum_{j=1}^{N_r} g_k(x_j; \mathbf{w}) = M_k, \quad \forall k \in \{1,...,K\}, \\
& 
\label{3}
\sum_{k=1}^{K} g_k(x_i; \mathbf{w}) = 1, \forall i \in \{1,...,N_f\},\quad \sum_{k=1}^{K} g_k(x_j; \mathbf{w}) = 1, \forall j \in \{1,...,N_r\} \\
& \label{4}
g_k(x_i; \mathbf{w}) \in [0, 1],  \forall i \in \{1,...,N_f\}, \forall k, \quad g_k(x_j; \mathbf{w}) \in [0, 1], \forall j \in \{1,...,N_r\}, \forall k
\end{align}
}

\subsubsection{Convergence to the Unique Optimal Solution}
To address computational efficiency for large datasets, we develop an iterative algorithm based on coordinate descent applied to our constrained optimization problem.                                                                 
\begin{theorem}
The proposed iterative procedure converges to the unique optimal solution, provided that feasible initial conditions are used and the KL divergence remains finite for all feasible distributions.
\end{theorem}
\textit{Proof sketch:}
The KL divergence \( D_{\text{KL}}(p \| q) \) is strictly convex in $p$ when $q$ is fixed. Since our objective function is a sum of strictly convex functions, it is strictly convex overall. Combined with linear constraints, this yields a convex optimization problem with a unique global minimum. Our coordinate descent method maintains feasibility through closed-form updates, guaranteeing convergence to the global optimum. 


\subsection{Optimization Algorithm}
We solve the constrained optimization problem using coordinate descent with Lagrangian multipliers.


\subsubsection{Lagrangian  Formulation}
To handle the constraints, we introduce Lagrange multipliers $\alpha_k$ for each class $k$:

\begin{equation}
\begin{split}
\mathcal{L}(f,\alpha)
= &\sum_{i=1}^{N_f} D_{\text{KL}}\!\big(g(x_i;\mathbf{w}) \,\|\, f(x_i;\mathbf{w})\big)
  + \lambda \sum_{j=1}^{N_r} D_{\text{KL}}\!\big(g(x_j;\mathbf{w}) \,\|\, f(x_j;\mathbf{w})\big) \\
&\quad + \sum_{k=1}^K \alpha_k \!\left(
      \sum_{i=1}^{N_f} g_k(x_i;\mathbf{w})
    + \sum_{j=1}^{N_r} g_k(x_j;\mathbf{w})
    - M_k \right)
\end{split}
\end{equation}

\subsubsection{Coordinate descent updates}

Taking derivatives and applying KKT conditions yields the closed-form updates:

\textbf{Primal updates:}
\begin{equation}
g_k(x_i; \mathbf{w}) = \hat{g}_{i,k} \exp(-\alpha_k), \quad 
g_k(x_j; \mathbf{w}) = \hat{g}_{j,k} \exp(-\alpha_k/\lambda)
\end{equation}

\textbf{Dual updates:}
\begin{equation}
\alpha_k^{(t+1)} = \alpha_k^{(t)} + \eta \left( \sum_{i=1}^{N_f} g_k^{(t)}(x_i; \mathbf{w}) + \sum_{j=1}^{N_r} g_k^{(t)}(x_j; \mathbf{w}) - M_k \right)
\end{equation}

where $\eta > 0$ is the step size. The algorithm alternates between these updates until convergence.

\subsection{Weight Update via Backpropagation}

After obtaining optimal output distributions through the above optimization, we update the model weights to realize these target distributions. We define a loss function based on the KL divergence:

\begin{equation}
\mathcal{L}_{\text{weight}} = \sum_{i=1}^{N_f} D_{\text{KL}}(g^*(x_i) \| f(x_i; \mathbf{w})) + \sum_{j=1}^{N_r} D_{\text{KL}}(g^*(x_j) \| f(x_j; \mathbf{w}))
\end{equation}

where $g^*$ denotes the optimal distributions from our optimization. The weights are updated via gradient descent:

\begin{equation}
\mathbf{w}^{(t+1)} = \mathbf{w}^{(t)} - \gamma \nabla_{\mathbf{w}} \mathcal{L}_{\text{weight}}
\end{equation}

This ensures the model's outputs converge to the optimized distributions that achieve unlearning while maintaining natural probability patterns.

\section{Experiment}

\subsection{Datasets and Metrics}
In this study, we employ three datasets that were also used in prior research: CIFAR-10, CIFAR-100, and Lacuna-10. Lacuna-10 is a curated dataset formed by selecting data from 10 distinct classes, randomly chosen from the extensive VGG-Face2 dataset \citep{cao2018vggface2}. These selected classes each have a minimum of 500 samples, with the data further segmented into 400 training and 100 testing images per class. Lacuna-100 expands on this concept by selecting 100 classes with the same criteria. 

Our evaluation employs multiple metrics to comprehensively assess unlearning performance. We measure the model's error rate (defined as $100\% - \text{accuracy}$) on three sets: the forget set to verify successful unlearning, the retain set to evaluate memory preservation, and the test set to assess generalization ability. For privacy protection tasks, we additionally evaluate the model's resistance to membership inference attacks. We also introduce a metric that measures the KL divergence between the output distributions of the unlearned and retrained models on forget set inputs, evaluating how closely the unlearned model approximates ideal retraining behavior.

\subsection{Implementation details}
To facilitate comprehensive comparison with the performance of other models, we follow the setup in \citep{kurmanji2024towards}. We establish two experimental conditions: small-scale and large-scale. The small-scale setting, referred to as CIFAR-5/Lacuna-5, involves a subset of 5 classes from each dataset, comprising 100 training, 25 validation, and 100 testing samples per class. Notably, the forget set includes 25 samples from the initial class, accounting for 5\% of the dataset. Conversely, the large-scale setting encompasses all classes from both CIFAR-10 and Lacuna-10, providing a broader spectrum for analysis. In the large-scale scenario, we will explore both class unlearning and selective unlearning. For class unlearning, we define the forget set as the entirety of the training set for class 5, which constitutes 10\% of the data. In the selective unlearning scenario, we aim to forget 100 examples from class 5, representing 0.25\% of CIFAR-10 and 2\% of Lacuna-10.

To align with precedents in the field, our experiments are conducted using two architectures: ResNet-18 and ALL-CNN \citep{springenberg2014striving}. The baseline model is pretrained on the CIFAR-100 and Lacuna-100 datasets for initial weight setting. Additionally, $\lambda$ is be set to a default value of 1 in the following experiments. More details of hyperparameters are are shown in Appendix~\ref{experiment}.

\subsection{Baseline}
Our approach is benchmarked against the other unlearning methods and established baselines to highlight its efficacy:
\textbf{Retrain}: The model is trained solely on the retain set $\mathcal{D}_r$, considered the gold standard. However, this method is typically deemed impractical for real-world applications. \textbf{Original}: The baseline model is trained on the complete dataset $\mathcal{D}$, without any modifications for data forgetting. \textbf{Finetuning}: The original model is fine-tuned on the retain set $\mathcal{D}_r$, incorporating no specific forgetting mechanism. $\textbf{NegGrad}+$ \citep{kodge2023deep}: An innovative method that applies gradient ascent to the forget set and gradient descent to the retain set over 500 iterations. \textbf{Fisher Forgetting} \citep{golatkar2020eternal}: Adjusts the model's weights to effectively ``unlearn" the data meant to be forgotten, simulating a scenario where the model was never exposed to this data. \textbf{NTK Forgetting} \citep{doan2021theoretical}: Employs novel techniques like PCA-OGD to minimize forgetting by orthogonally projecting onto principal directions, preserving data structure integrity. \textbf{CF-k, EU-k} \citep{goel2022towards}: These methods focus on the model's last k layers. ``Exact-unlearning" (EU-k) re-trains these layers from scratch, while ``Catastrophic Forgetting" (CF-k) fine-tunes them on the retain set $\mathcal{D}_r$.
\textbf{SCRUB}~\citep{kurmanji2024towards}: Introduces a novel training objective and has demonstrated superior performance in prior metrics.

\subsection{Privacy Protection}

\begin{table*}[t]
\scriptsize  
  \caption{KL divergence between output distributions of unlearned and retrained models on forget set inputs. A lower KL divergence indicates closer alignment with the retrained model's output distribution, providing better privacy protection. }
  \label{kl_forget}
  \centering
  \begin{tabular}{l|c|c}
    \toprule
Task  & KL(AdaProb$\|$Retrain) ($\downarrow$) & KL(SCRUB$\|$Retrain) ($\downarrow$) \\
    \midrule
ResNet on Lacuna-5& 1.35& 3.65\\
ResNet on Lacuna-10 & 5.76 & 5.88\\
ResNet on CIFAR-5 & 2.56 & 3.01\\
ResNet on CIFAR-10 & 7.89 & 7.79  \\
ALLCNN on Lacuna-5 & 2.02 & 2.23 \\
  \bottomrule
  \end{tabular}
\end{table*}

For privacy protection, our goal is to ensure that the forget error remains close to that of retraining to avoid leakage of privacy. We evaluate privacy protection through membership inference attacks, which is adopted from the approach outlined by~\citet{kurmanji2024towards}. Specifically, we train a binary classifier (the ``attacker") using the losses of the unlearned model on both the forget and test examples, with the objective of classifying instances as either ``in" (forget) or ``out" (test). The attacker then predicts labels for held-out losses—losses that were not used during training—balanced between the forget and test sets. A successful defense is indicated by an attacker's accuracy of 50\%, signifying that the attacker is unable to distinguish between the two sets, demonstrating the effectiveness of the unlearning method.

According to Table~\ref{cifar10_resnet_privacy}, AdaProb's forget error is very close to that of retraining, particularly in the Lacuna-10 experiment, where it is the closest match. In the membership inference attack experiment, shown in Table~\ref{MIA}, AdaProb consistently achieves nearly 50\% accuracy, indicating strong privacy preservation. This demonstrates that, with the refinement of pseudo-probabilities, the model can maintain the original distribution while effectively forgetting the designated forget set. 

We conducted additional experiments on privacy protection tasks shown in Table~\ref{cifar10_resnet_privacy}, evaluating forget, retain, and test set errors. Our results show that AdaProb achieves performance nearly identical to the retrained model across all sets, providing strong evidence of effective privacy protection.

\begin{table*}[h]
\scriptsize  
  \caption{Unlearning results with ALL-CNN for the privacy protection task.}
  \label{cifar10_resnet_privacy}
  \centering
  \begin{tabular}{l|ccc|ccc}
    \toprule
    & \multicolumn{3}{c}{CIFAR-10} & \multicolumn{3}{c}{Lacuna-10} \\
Model  & Test error  & Retain error  & Forget error  & Test error  & Retain error  & Forget error  \\
    \midrule
    Retrain  & 16.71 & 0.00 & 26.67 &1.50& 0.00 & 0.33  \\
Original & 16.71 &0.00 & 0.00 &1.57& 0.00 & 0.00  \\
Finetune & 16.86&0.00 & 0.00 & 1.40&0.00& 0.00  \\
NegGrad+  & 21.65 &4.54 & 47.00 &3.60&0.87 & 14.33  \\
CF-k &16.82 &0.00 & 0.00 &1.57& 0.00 & 0.00 \\
EU-k &18.44 &0.32 & 0.33 & 3.90&0.76 & 0.00 \\
Bad-T & 22.43& 10.13 & 1.67 &4.90& 0.67 & 1.34  \\
SCRUB & 17.01 & 0.00 & 33.00 &1.67& 0.00 & 0.00 \\
SCRUB+R & 16.88 & 0.00 & 26.33 &1.67& 0.00 & 0.00 \\
\textbf{AdaProb} & 18.05& 0.00 &25.35 & 1.05 & 0.00 & 0.05\\
  \bottomrule
  \end{tabular}
\end{table*}

\begin{figure}[h]
\begin{center}
    \includegraphics[width=1.0\textwidth]{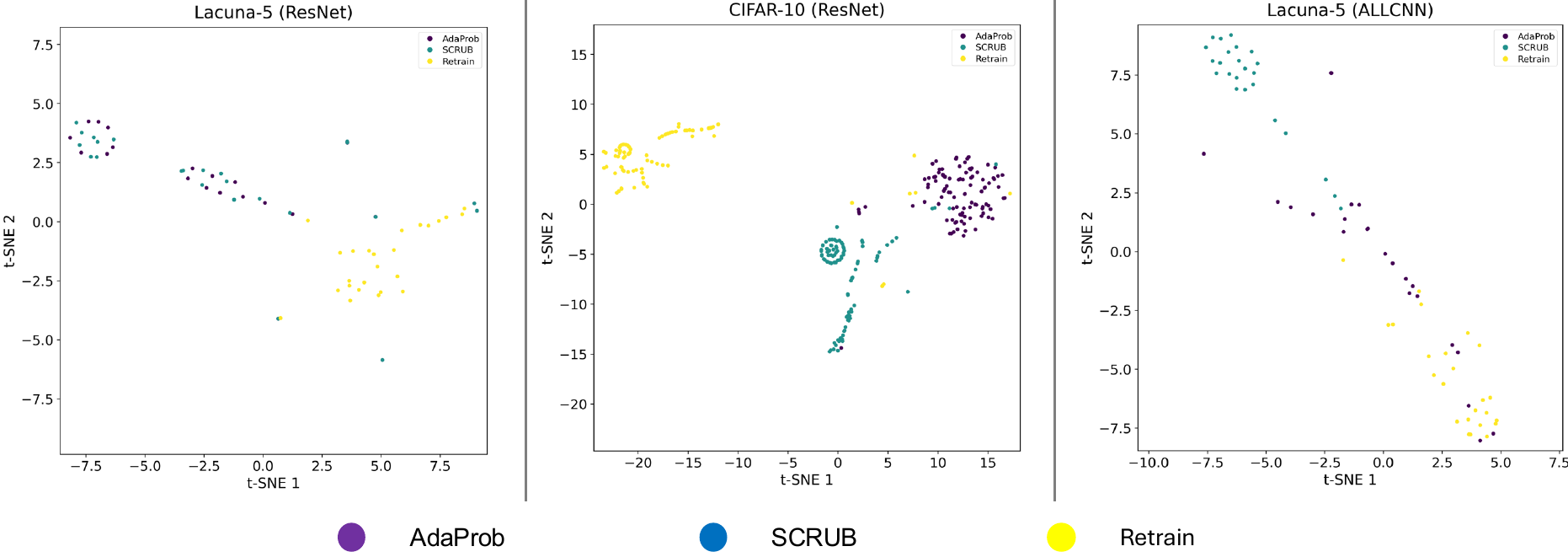} 
    \caption{t-SNE map of output distributions on the forget set for Retrain, SCRUB, and our unlearned model. Points closer to the Retrain cluster indicate stronger privacy (better unlearning).}
    \label{t_sne}
\end{center}
\end{figure}

We evaluate the similarity between unlearned and retrained models by measuring the KL divergence between their output distributions on forget set inputs, using SCRUB as a baseline. As illustrated in the t-SNE visualization in Figure~\ref{t_sne}, the output probabilities of AdaProb (purple points) cluster more closely to those of the retrained model (yellow points) compared to SCRUB (blue points) in ALLCNN trained on Lacuna-5. In other settings, our method produces output distributions comparable to SCRUB. Table~\ref{kl_forget} presents KL divergence values that support this observation, showing that AdaProb achieves output distributions closer to the retrained model in certain cases, while matching SCRUB's performance in others. This demonstrates that AdaProb consistently approximates the behavior of a model that was never exposed to the forgotten data, performing at least as well as SCRUB across different scenarios. When considering the significantly reduced computation time and enhanced resistance to membership inference attacks, AdaProb emerges as the superior method. Additionally, Table~\ref{kl} reports the KL divergence values between output distributions on the test set, further validating that our approach has better privacy protection.

\begin{figure}[h]
\begin{center}
    \includegraphics[width=1.0\textwidth]{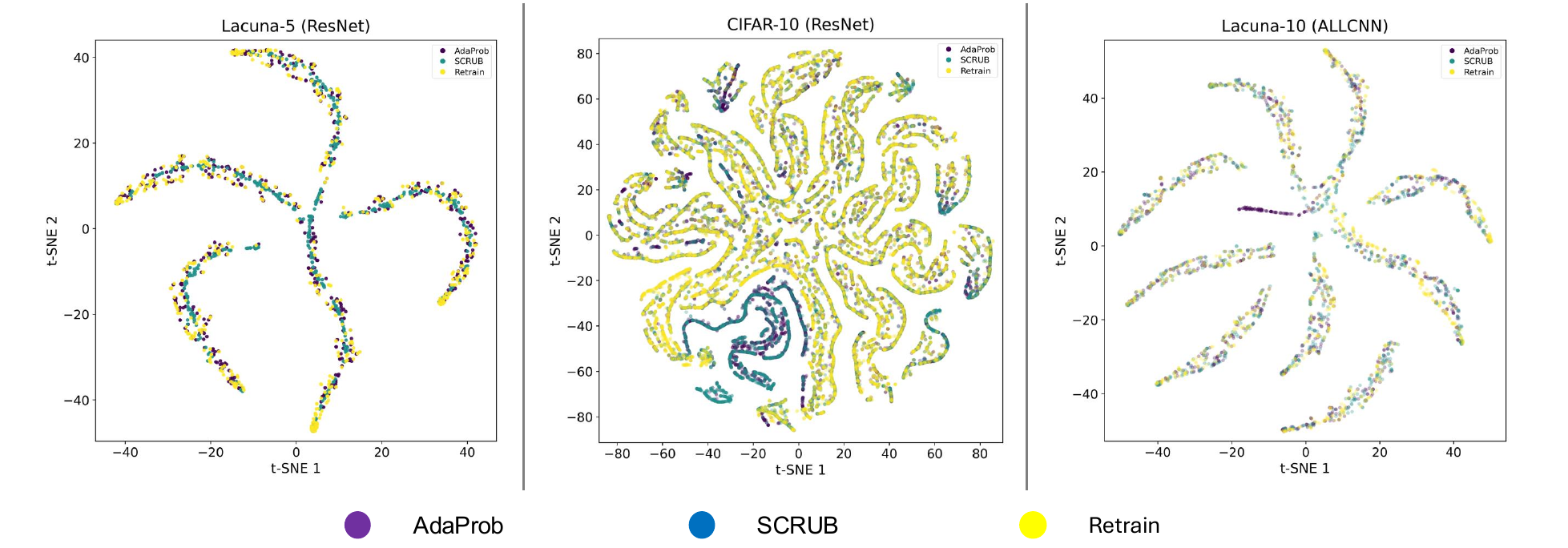} 
    \caption{t-SNE map of output distributions on the test set for Retrain, SCRUB, and our unlearned model. Points closer to the Retrain cluster indicate stronger privacy (better unlearning).}
    \label{t_sne_test}
\end{center}
\end{figure}

In addition to calculating KL divergence on the forget set, we investigated the model's generalization ability through additional experiments on the test set. The results in Table ~\ref{kl} show that AdaProb achieves lower KL divergence compared to SCRUB when measured against the retrained model, indicating that AdaProb produces an unlearned model that more closely resembles the ideal retraining baseline. Also, Figure~\ref{t_sne_test} use t-SNE map to helps visualize the output distribution of retrain, SCRUB, and AdaProb. 

\begin{table*}[t]
\scriptsize  
  \caption{KL divergence between output distributions of unlearned and retrained models on test set inputs. A lower KL divergence indicates closer alignment with the retrained model's output distribution. }
  \label{kl}
  \centering
  \begin{tabular}{l|c|c}
    \toprule
Task  & KL(AdaProb$\|$Retrain) ($\downarrow$) & KL(SCRUB$\|$Retrain) ($\downarrow$) \\
    \midrule
ResNet on Lacuna-5& 0.044& 0.21\\
ResNet on Lacuna-10 & 0.76 & 0.88\\
ResNet on CIFAR-5 & 0.056 & 0.11\\
ResNet on CIFAR-10 & 0.18 & 0.20 \\
ALLCNN on Lacuna-5 & 0.10 & 0.09 \\
ALLCNN on Lacuna-10 & 0.22 & 0.23\\
  \bottomrule
  \end{tabular}
\end{table*}

\begin{table*}[h]
\scriptsize  
  \caption{Membership inference attack results with ResNet-18 and ALL-CNN in large-scale unlearning. The closer the result is to 50\%, the better the performance. }
  \label{MIA}
  \centering
  \begin{tabular}{l|cc|cc|cc|cc}
    \toprule
& \multicolumn{4}{c}{ResNet} &\multicolumn{4}{c}{ALL-CNN} \\
    & \multicolumn{2}{c}{Class} & \multicolumn{2}{c}{Selective} &\multicolumn{2}{c}{Class} & \multicolumn{2}{c}{Selective} \\
Model  & mean  & std & mean  & std & mean  & std & mean  & std \\
    \midrule
    Retrain  & 49.33 &1.67 &54.00 &1.63 &55.00 &4.00& 48.73& 0.24  \\
Original & 71.10& 0.67 &65.33 &0.47& 66.50& 0.50& 71.40& 0.70  \\
Finetune & 75.57 &0.69 &64.00& 0.82& 68.00& 1.00& 74.97& 1.27 \\
NegGrad+  & 69.57 &1.19& 66.67 &1.70& 72.00& 0.00& 70.03& 1.92  \\
CF-k &75.73& 0.34& 65.00& 0.00& 69.00& 2.00& 72.93& 1.06 \\
EU-k &54.20 &2.27& 53.00& 3.27& 66.50& 3.50& 51.60& 1.22 \\
Bad-T & 54.00& 1.10& 59.67& 4.19& 63.40& 1.20& 77.67& 4.11  \\
SCRUB & 52.20& 1.71& 78.00& 2.45& 52.00& 0.00& 54.30& 2.24 \\
SCRUB+R & 52.20& 1.71& 58.67& 1.89& $\mathbf{52.00}$& 0.00& 54.30& 2.24\\
\textbf{AdaProb} & $\mathbf{51.00}$ & 1.05 &  $\mathbf{58.00}$  &0.93     &   54.00  &  0.70  &  $\mathbf{50.00}$   &0.40\\

  \bottomrule
  \end{tabular}
\end{table*}

\subsection{Computational efficiency}
We compare the time required for SCRUB \citep{kurmanji2024towards}, retraining, and our method, with all experiments conducted on an NVIDIA RTX-4090. Time is recorded over 5 runs, and we report both the mean and the standard error. In Figure~\ref{time_fig}, we present the time required for the tasks using the ResNet-18 model and selective unlearning using ALL-CNN. Compared to other methods, AdaProb significantly reduces computation time, cutting it to less than half of what is required by SCRUB. The results further emphasize the high effectiveness of the optimization approach and the use of pseudo-probabilities to fine-tune the model weights.
\begin{figure}[h!]
\begin{center}
    \includegraphics[width=0.9\textwidth]{figure.pdf}  
    \caption{Time needed for the unlearning method (measured over 5 runs)}
    \label{time_fig}
\end{center}
\end{figure}

\subsection{Ablation studies}
We conduct two further ablation studies. First, in the optimization objective function (\ref{1}), the value of 
$\lambda$ was set to 1 in all previous experiments. In \autoref{tab:ablation}, we explore the impact of varying 
$\lambda$ on the retain and forget errors in a small-scale unlearning experiment on CIFAR-5 with ResNet. As 
$\lambda$ increases, more weight is assigned to the retain set, resulting in a decrease in retain error from 0.21\% to 0\%. However, this reduction comes at a significant cost to the forget error.

Second, we investigate our method in a larger setting, using the CIFAR-100 dataset with one class unlearning. Our method demonstrated very good performance. Using the ResNet architecture, SCRUB achieved a forget error of 5.19\% and a retain error of 0.00\%. In contrast, our method achieved a retrain error of 0.00\% and a forget error of 98.25\%.

\begin{table*}[t!]
\scriptsize  
  \caption{The retain error and forget error with varying 
$\lambda$ values were evaluated in a small-scale unlearning experiment on CIFAR-5 using ResNet.}
  \label{tab:ablation}
  \centering
  \begin{tabular}{l|cc|cc|cc|cc}
    \toprule
    & \multicolumn{2}{c}{$\lambda = 1$}& \multicolumn{2}{c}{$\lambda = 2$} &\multicolumn{2}{c}{$\lambda = 3$}  &\multicolumn{2}{c}{$\lambda = 4$} \\

Model  & Retain error & Forget error & Retain error & Forget error  & Retain error & Forget error& Retain error & Forget error\\
    \midrule
\textbf{AdaProb} & 0.21 & 80.00 &  0.00  &56.00     &   0.00 &  23.00  &  0.00   &30.00\\

  \bottomrule
  \end{tabular}
\end{table*}

\section{Conclusion}

This research introduces a novel approach to machine unlearning, presenting an optimization framework that refines output probability distributions within deep learning models. Our method excels in striking an optimal balance between forgetting effectiveness and preserving model performance. Additionally, it demonstrates superior resilience against membership inference attacks. Empirical results across diverse datasets and model architectures, including CIFAR-10 and Lacuna-10 with ResNet and ALL-CNN, highlight the superiority of our approach over existing methods.

Furthermore, the operational flexibility, theoretical insights, and high computational efficiency of our approach provide a solid foundation for further developments. However, we acknowledge certain limitations. Our current method is limited to addressing unlearning in classification tasks and may encounter convergence issues during the optimization process.
Additionally, the approach is restricted to supervised learning settings and does not extend to unsupervised tasks at this stage.
Future work will focus on extending the method to various models, including large language models, and broadening its applicability beyond classification tasks.

\section*{Acknowledgment}

This work was supported in part by National Science Foundation (NSF) under grant OAC-23-19742. The views and
conclusions contained herein are those of the authors and
should not be interpreted as necessarily representing the official policies or endorsements, either expressed or implied, of
NSF.

\bibliography{iclr2026_conference}
\bibliographystyle{iclr2026_conference}

\appendix
\section{Experiment details}
\label{experiment}
This section presents the hyperparameters used in our experiments. Table\ref{pretrain} details the pretraining configuration, while Table\ref{train} specifies the training parameters. 
\begin{table*}[h]
\scriptsize  
  \caption{Hyperparameter for pretrained models}
  \label{pretrain}
  \centering
  \begin{tabular}{l|c|c}
    \toprule
    Model & filter & learning rate\\
    \midrule
ALLCNN & 0.4 & 0.1 \\
ResNet & 1.0 & 0.1 \\
  \bottomrule
  \end{tabular}
\end{table*}

\begin{table*}[h]
\scriptsize  
  \caption{Hyperparameter for training models}
  \label{train}
  \centering
  \begin{tabular}{l|c|c|c|c|c|c}
    \toprule
    Model & filter & learning rate & weight decay & batch size & epochs & seed\\
    \midrule
ResNet (CIFAR5) & 0.4 & 0.001 & 0.1 & 128 & 31 & 3 \\
ALLCNN (CIFAR5) & 1.0 & 0.001 & 0.1 & 128 & 31 & 3\\
ResNet (CIFAR5) & 0.4 & 0.001 & 0.1 & 128 & 31 & 3 \\
ALLCNN (Lacuna5) & 1.0 & 0.001 & 0.1 & 128 & 31 & 3\\
ResNet (CIFAR10) & 1.0 & 0.01 & 5e-4 & 128 & 26 & 1 \\
ALLCNN (CIFAR10) & 1.0 & 0.01 & 5e-4 & 128 & 26 & 1\\
ResNet (Lacuna10) & 1.0 & 0.01 & 5e-4 & 128 & 26 & 1 \\
ALLCNN (Lacuna10) & 1.0 & 0.01 & 5e-4 & 128 & 26 & 1\\
  \bottomrule
  \end{tabular}
\end{table*}
\end{document}